\documentclass[11pt]{article}

\usepackage[utf8]{inputenc}
\usepackage[T1]{fontenc}
\usepackage[margin=1in]{geometry}
\usepackage{amsmath,amssymb}
\usepackage{booktabs}
\usepackage{graphicx}
\usepackage{xcolor}
\usepackage{xspace}
\usepackage{authblk}
\usepackage[numbers,round]{natbib}
\usepackage{caption}
\usepackage[hidelinks]{hyperref}

\graphicspath{{figures/}}

\newcommand{\dirfrac}{\texttt{dir\_frac}\xspace}
\newcommand{\Dhead}{\ensuremath{D_{\text{head}}}\xspace}
\newcommand{\contentpos}{\texttt{content\_pos\_frac}\xspace}
\newcommand{\imagmass}{\texttt{imag\_mass}\xspace}
\newcommand{\ropeimag}{\texttt{rope\_imag\_frac}\xspace}
\newcommand{\selfmatch}{\texttt{self\_match}\xspace}
\newcommand{\M}{\ensuremath{M}\xspace}
\newcommand{\MS}{\ensuremath{M_S}\xspace}
\newcommand{\MA}{\ensuremath{M_A}\xspace}
\newcommand{\Mt}{\ensuremath{M_t}\xspace}
\newcommand{\Wq}{\ensuremath{W_q}}
\newcommand{\Wk}{\ensuremath{W_k}}
\newcommand{\Rope}{RoPE\xspace}

\title{\textbf{Fingerprint, Not Blueprint:\\
How Positional Schemes Set the Default Spectral Algebra of Attention}}
\author[1]{Li Hengyu}
\affil[1]{Institute for Solid State Physics, The University of Tokyo\\
\texttt{lihengyu@issp.u-tokyo.ac.jp}}
\date{}

\begin{document}
\maketitle

\begin{abstract}
The pre-softmax score of an attention head is a bilinear form $\mathrm{score}(i,j)=x_i^\top \M x_j$ in a learned
operator $\M = \Wq^\top \Wk$. Because \M is generally non-symmetric, hence non-normal, it has a complex
eigenspectrum and non-orthogonal eigenvectors, the regime where non-Hermitian and random-matrix tools apply. We ask
what this spectrum encodes, at three levels for previous-token and induction circuits. Statically, across seven
pretrained models spanning three positional schemes, the strongest previous-token heads are spectrally rotational
under \Rope and non-rotational, or content-like, where position enters outside QK (learned-absolute and ALiBi); the
model-level separation is perfect at every top-$k$ examined (exact permutation $p{=}0.029$), and zeroing the
per-frequency \Rope phase $\mathrm{Im}(\Mt)$ eliminates induction on a pre-identified previous-token head in all
three \Rope models. Dynamically, over public Pythia checkpoints every head originates at the random-matrix
(Ginibre) null; the rotational signature emerges with the behavior, not before it, and the population-median
suppression that yields the final profile follows circuit formation, so the profile is a consolidated fingerprint,
not a precursor. Causally, and at toy scale, no spectral channel is necessary: constrained two-layer training
reroutes around every ban with capability intact, albeit at a significant formation delay (four pre-registered
contrasts, $q_{\mathrm{BH}}{\le}0.016$). The cost structure exposes each scheme's default: imposing symmetry slows
learned-absolute models by a factor of $2.9$, whereas a \Rope head with a fully symmetric static \M still routes
directionally via the phase channel, impossible under absolute positions. Within the settings examined, the
positional scheme sets the default spectral algebra of an attention head's solution: a fingerprint sculpted after
function, not a hard constraint upon it.
\end{abstract}

\section{Introduction}
For a single attention head with norm-processed residual token $x\in\mathbb{R}^d$ and projections $\Wq,\Wk$, the
pre-softmax score is a bilinear form in one operator,
\begin{equation}
\mathrm{score}(i,j) = q_i^\top k_j = x_i^\top(\Wq^\top \Wk)x_j = x_i^\top \M x_j,\qquad \M=\Wq^\top \Wk \in\mathbb{R}^{d\times d}.
\end{equation}
\M is the natural, gauge-invariant object of study: the reparametrization $\Wq\!\to\!\Wq G,\ \Wk\!\to\!\Wk G^{-\top}$
leaves \M and every observable unchanged, so raw $\Wq/\Wk$ entries are gauge artifacts while functions of \M are
physical \citep{elhage2021framework}. \M is low rank ($\mathrm{rank}\le d_k$) and, crucially, \emph{generally
non-symmetric}---so it has a complex eigenspectrum.

A growing literature reads non-normal/complex spectral structure into transformers: Schur decompositions of the
residual-stream Jacobian \citep{fernando2026dynamics}, ``non-Hermitian'' operator-theoretic framings
\citep{chang2026dyson}, and symmetric/antisymmetric decompositions of the QK matrix \citep{saponati2025}; concurrent
work catalogues per-head complex eigenvalues of the attention interaction and ablates its skew/symmetric channels
against corpus perplexity \citep{jamil2026routing}. Yet the QK
operator's complex spectrum remains \emph{behaviorally unanchored} --- no random-matrix nulls, no head-function
taxonomy, no head-resolved causal tests, no positional-scheme contrasts --- and the field's discipline demands we ask
not whether the structure \emph{exists} but whether it \emph{does measurable work}. Concretely: does the
complex-eigenvalue view of \M predict head behavior that the plain symmetric/antisymmetric split does not already
capture?

We answer this at three levels: \emph{statics} across seven pretrained models --- GPT-2 small, OPT-1.3B, and
GPT-Neo-1.3B (learned-absolute positions), BLOOM-1b1 (ALiBi: position enters as a score bias, leaving QK positionally
unconstrained), Pythia-410m/1.4B ($25\%$ \Rope), and Llama-3-8B ($100\%$ \Rope, RMSNorm, GQA); \emph{dynamics} across
the public Pythia training checkpoints; and \emph{interventions} in constrained-from-scratch training runs. Our
contributions:
\begin{enumerate}\itemsep2pt
\item \textbf{A matched-null spectral framework for the QK operator} (\S\ref{sec:method}): four decompositions of the
norm-folded \M, a primary directionality metric \Dhead and its plain-split baseline \dirfrac, and---essential---a
\emph{random-orientation (Ginibre) null} against which directionality must be read (a random low-rank \M already has
$\dirfrac\approx1/\sqrt2$).
\item \textbf{Descriptive}: spectral directionality separates head function across all seven models
(\S\ref{sec:descriptive}), extending \citet{saponati2025} from training objective to head function.
\item \textbf{Causal anatomy} on GPT-2 (\S\ref{sec:causal}): the symmetric part \MS is $6.7\times$ more load-bearing
than the antisymmetric \MA; \MA is causally critical only for the canonical induction/previous-token circuit and inert
elsewhere. On GPT-2 the complex refinement \Dhead does \emph{not} beat \dirfrac.
\item \textbf{The architecture-conditional signature} (\S\ref{sec:reversal}): the strongest previous-token heads
are spectrally rotational under \Rope (top-quintile \Dhead) and non-rotational under learned-absolute \emph{and}
ALiBi positions (bottom-quartile, four models); the model-level separation is perfect at every top-$k$ tested
(exact permutation $p{=}0.029$; head-level $p{=}2.5{\times}10^{-5}$, descriptive); $d_k$ held fixed across schemes.
Aggregate partials are shown to be bulk-contaminated and unreliable for this contrast.
\item \textbf{Mechanism} (\S\ref{sec:mechanism}): a per-frequency complex decomposition ties \Dhead to the \Rope
rotational phase $\mathrm{Im}(\Mt)$; adding \Rope-phase features as controls attenuates \Dhead's advantage by
${\sim}47$--$61\%$ ($25\%$ \Rope) and ${\sim}70$--$98\%$ ($100\%$ \Rope), estimator-dependent.
\item \textbf{Causal confirmation} (\S\ref{sec:ablation}): ablating $\mathrm{Im}(\Mt)$ symmetrizes the relative-position
kernel and destroys induction on exactly the flagged heads.
\item \textbf{A checkpoint natural history} (\S\ref{sec:dynamics}): on Pythia-410m/160m (22 public checkpoints each),
all heads are born at the Ginibre null; the rotational signature locks in \emph{with} circuit formation (not before
it), and the population-median suppression that produces the static profile follows formation --- the profile is a
consolidated fingerprint, not a precursor.
\item \textbf{Constrained-training interventions} (\S\ref{sec:intervention}): a $\{$APE, \Rope$\}\times\{$free,
sym-\M, $\mathrm{Im}(\Mt)$-suppressed$\}$ grid ($n{=}5$ seeds) shows no spectral channel is \emph{necessary}
(every arm reroutes) while every constraint carries a significant search cost that reveals the scheme's default
solution, and dissociates the static-antisymmetry and \Rope-phase channels at the weight level.
\end{enumerate}
One sentence unifies the three levels: \emph{the positional scheme sets the default spectral algebra of attention's
solutions --- a fingerprint sculpted after function, an economy of search rather than a hard constraint.}
We report the negatives (GPT-2 decorativeness; \MA's inertness away from a few heads) as plainly as the positives.

\section{Related work}
\textbf{QK circuits and weight-space interpretability.} \citet{elhage2021framework} frame $\M{=}\Wq^\top\Wk$ and
$W_{OV}$ as the two low-rank circuits and use \emph{OV} eigenvalue positivity as a copying statistic; they perform no
QK spectral analysis. \citet{millidge2022svd} find weight-SVD interpretable but report it \emph{fails} on QK---motivating
an orientation-recovering (Schur/complex) view. Our object and the OV/QK distinction follow this line; our contribution
is the QK-side, imaginary-axis, behavior-anchored analog.

\textbf{Symmetric/antisymmetric QK.} \citet{saponati2025} decompose $W_{qk}{=}M_s{+}M_n$ and define a Frobenius
symmetry score, anchoring it to the \emph{training objective}. Our \dirfrac is a monotone transform of their score
($s = 1-2\,\dirfrac^2$); we anchor the split to \emph{head function} and add the complex/Schur refinement and its
causal test.

\textbf{Non-normal / non-Hermitian transformers.} Concurrently with this work, \citet{jamil2026routing} decompose
the per-head attention interaction into skew (``routing'') and symmetric (``filtering'') channels across $1{,}776$
heads of five pretrained models, report $\max\mathrm{Re}\,\lambda>0$ for every head, find by all-heads truncation
surgery on GPT-2 Large that the symmetric channel is by far the more load-bearing (convergent, at the aggregate
level, with our \S\ref{sec:causal}), and train a stability-constrained skew-minus-diagonal attention from scratch.
Their analysis stops at norms, ranks, and stability --- no behavioral taxonomy, no random-matrix null, no
head-resolved causal test, no positional-scheme contrast, and their weight-level routing--filtering ratio is (like
Saponati's score) a monotone transform of \dirfrac --- so it cannot ask whether the complex spectrum does work
\emph{beyond} the plain split; that anchoring is our contribution. \citet{fernando2026dynamics} Schur-decompose the residual-stream
\emph{Jacobian} and show learned non-normality is functionally central---but on a different operator, without
pseudospectra, and anchored to rank/propagation, not head function. \citet{chang2026dyson} labels $\Wq^\top\Wk$
non-Hermitian in a Dyson-series theory with no computation. We differ by analyzing the QK operator's complex spectrum
empirically and anchoring it to head behavior.

\textbf{RoPE and positional heads.} \Rope \citep{su2021roformer} injects $e^{im\theta_t}$ phase,
$\theta_t=\mathrm{base}^{-2t/d}$. \citet{barbero2025rope} and \citet{urrutia2025decoupling} show positional heads use
high \Rope frequencies. We connect a \emph{weight-space} directionality scalar to per-head \Rope-frequency usage and to
head function, which they do not.

\textbf{Circuit formation over training.} \citet{olsson2022induction} established the induction phase change;
\citet{tigges2024llm} track circuits across all Pythia checkpoints behaviorally (induction at ${\sim}2{\times}10^9$
tokens); toy-model studies dissect formation \citep{bietti2023birth,reddy2024mechanistic,singh2024needs}; and
\citet{chen2024sudden} regularize an interpretable attention property during MLM pretraining --- the template our
intervention ports to autoregressive weight space. \citet{saponati2025} track their symmetry score during training
(per-layer, own encoder/decoder runs). None of these tracks a weight-space spectral quantity per head across public
checkpoints against the phase change (\S\ref{sec:dynamics}), and none constrains QK spectral structure during
autoregressive training (\S\ref{sec:intervention}).

\textbf{Constrained-attention training.} Hard symmetric (shared-QK) causal LMs exist --- Reformer
\citep{kitaev2020reformer} and recent systematic sharing studies --- and symmetric dot-product BERT trains well
\citep{courtois2024symmetric}, but none analyzes circuits under the constraint, and the soft, decomposed versions
(penalize only \MA, or only $\mathrm{Im}(\Mt)$) are new here; concurrent observational work crosses positional
schemes with induction formation in toy models \citep{huang2026dissecting}.

\textbf{Head taxonomy.} We use standard detectors---previous-token, duplicate, induction \citep{olsson2022induction}---and
IOI \citep{wang2022ioi} head classes for validation.

\section{Method}\label{sec:method}
\textbf{Object and folding.} We use TransformerLens with \texttt{fold\_ln=True} to absorb the norm gain into $\Wq,\Wk$,
and analyze the norm-folded operator $M_{\text{eff}}=C\M C$ for LayerNorm (centering $C=I-\tfrac1d\mathbf{1}\mathbf{1}^\top$)
and $M_{\text{eff}}=\M$ for RMSNorm. All spectral work is float64. We verify the convention by reconstructing the
model's own pre-softmax scores from \M and matching to $<10^{-4}$ (learned-absolute) or, for \Rope models where position
cancels only at $i{=}j$, matching the diagonal (bf16 models match to bf16 precision; the fast metric path is exact,
$\dirfrac\ \Delta{=}10^{-16}$). A random $G\in GL(d_k)$ leaves \M, its eigenvalues, and \Dhead invariant to ${\sim}10^{-14}$
while changing raw column norms---confirming gauge hygiene.

{\sloppy\textbf{Decompositions.} Per head we compute the SVD, the symmetric/antisymmetric split $\M{=}\MS{+}\MA$,
the real Schur form ($2\times2$ blocks $\to$ rotation angles $\theta$), and the complex eigendecomposition (with
eigenvector condition number as a non-normality flag).\par}

\textbf{Metrics.}
\begin{align}
\dirfrac &= \|\MA\|_F/\|\M\|_F && \text{(plain antisymmetric magnitude; Saponati's score)}\\
\Dhead &= \textstyle\sum_r |\mathrm{Im}\,\lambda_r| \big/ \sum_r |\lambda_r| && \text{(primary complex/Schur directionality)}\\
\contentpos &= \textstyle\sum \lambda^+ \big/ \sum |\lambda| \ \text{of}\ \MS && \text{(signed content axis; QK-side echo of OV positivity)}
\end{align}
plus $\text{henrici}=\sqrt{\|\M\|_F^2-\sum|\lambda|^2}/\|\M\|_F$ (departure from normality) and \selfmatch (diagonal
dominance of $W_E \M W_E^\top$).

\textbf{The random-orientation null (essential).} A random low-rank \M is already highly ``directional'': for
independent Gaussian $\Wq,\Wk$ the symmetric and antisymmetric parts get equal expected Frobenius mass, so
$\dirfrac_{\text{null}}\approx 1/\sqrt2\approx0.707$ and \Dhead's null is likewise high. We therefore report every
metric relative to a \emph{matched-$\Sigma$ random-orientation null} (same singular values, random left/right frames),
computed exactly in the rank-$k$ core. Empirically $\dirfrac_{\text{null}}{=}0.707$, $D_{\text{head,null}}{=}0.608$;
$141/144$ GPT-2 heads deviate ($|z|{>}2$ on \dirfrac or \Dhead; \dirfrac alone: $140/144$). Nulls are computed
exhaustively for GPT-2; the cross-model analyses of \S\ref{sec:reversal} use raw metrics --- on GPT-2,
null-referenced metrics give an indistinguishable partial ($-0.204$ vs $-0.206$), so this choice is not load-bearing.

\textbf{Models \& corpus.} GPT-2 small (144 heads), OPT-1.3B (768), GPT-Neo-1.3B (384), BLOOM-1b1 (384; ALiBi),
Pythia-410m/1.4B (384), Llama-3-8B (1024; GQA expanded, RMSNorm, bf16). Behavioral runs use a $1\mathrm{k}\times128$ slice of \texttt{NeelNanda/pile-10k}; detectors use repeated-random
sequences. All on one A100.

\textbf{Checkpoint and intervention pipelines.} For \S\ref{sec:dynamics} we load Pythia public checkpoints natively
(TransformerLens \texttt{checkpoint\_value}); the fused-QKV unpacking and folding conventions were verified against
the main-revision pipeline to machine precision (weight metrics at the final checkpoint match the independent
pipeline exactly, $\Delta{=}0$). For \S\ref{sec:intervention} we train 2-layer attention-only models from scratch
with soft spectral penalties; definitions and the crucial sym-\M\,$\neq$\,$\mathrm{Im}(\Mt)$ distinction are given
there. Pre-registration: the dynamics questions (Q1--Q3) and intervention predictions (P1--P4) were locked in the
project's intervention plan before any checkpoint was downloaded or any constrained run launched; we report the
falsified predictions as findings.

\section{Spectral directionality separates head function}\label{sec:descriptive}
\begin{figure}[t]\centering
\includegraphics[width=0.6\linewidth]{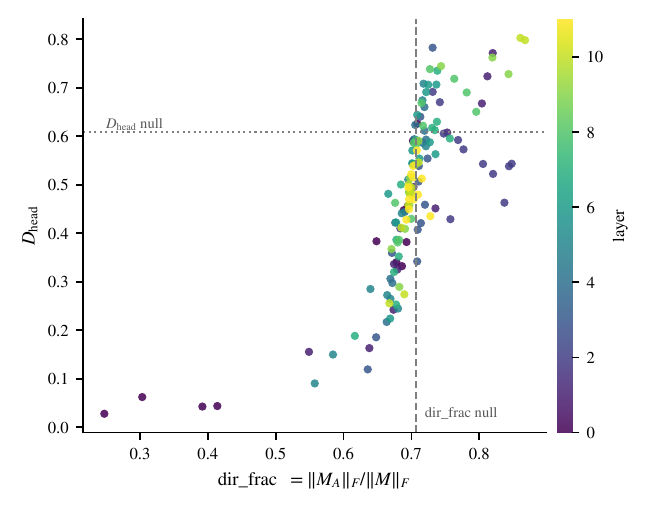}
\caption{The content--direction plane (GPT-2). Each point is a head; dashed lines mark the random-orientation null
$(\dirfrac_{\text{null}}{\approx}0.707,\ D_{\text{head,null}}{\approx}0.608)$. A content cluster (low on both, mostly
layer 0) sits below the null; directionality is read as deviation from it.}
\label{fig:plane}
\end{figure}

On GPT-2 ($n{=}144$, Spearman, BH-FDR $q{<}10^{-3}$), directionality maps onto head class (Table~\ref{tab:taxonomy}).
Previous-token heads are directional (high \dirfrac, low content positivity); duplicate/similarity heads are symmetric;
\textbf{induction heads read as content-like}. We note plainly that our preregistered hypothesis predicted the
opposite (induction heads directional); that prediction \emph{failed}. The two-head-circuit reading --- the induction
head's QK consumes \emph{composed content} written by a previous-token head, so its single-head \M is symmetric ---
was recorded in the analysis plan before these correlations were computed, and we verify it directly:
by the K-composition statistic (the Frobenius overlap of an induction head's QK operator with an earlier head's OV
write, ranked over all earlier heads), \textbf{the canonical previous-token head ranks \#1 for every induction head
tested} --- GPT-2's 4.11 for 5.1/5.5/6.9 (ranks 1/60, 1/60, 1/72) and Pythia-410m's 5.2 for 8.6/10.9/11.14 (1/128,
1/160, 1/176) --- confirming that induction QK is wired to consume the prev head's composed output. \textbf{Copying (OV) is orthogonal to every QK metric}, a clean control. The content
signature is legible: \selfmatch equals \contentpos at $r{=}0.96$; empirical pre-softmax score asymmetry on real
inputs tracks \dirfrac at $0.66$. This replicates on Pythia and, attenuated, on Llama (\contentpos$\to$prev $-0.30$
vs GPT-2's $-0.63$; \S\ref{sec:reversal}), establishing spectral directionality as an architecture-general
head-function signature and extending \citet{saponati2025} from objective to function (Fig.~\ref{fig:plane}).

\begin{table}[t]\centering\small
\begin{tabular}{lcccc}\toprule
metric & prev-token & duplicate & induction & copying (OV)\\\midrule
\dirfrac      & $+0.47$ & $-0.59$ & $-0.50$ & $0.06$ (ns)\\
\contentpos   & $-0.63$ & $+0.59$ & $+0.57$ & ns\\
\imagmass     & $+0.56$ & $-0.65$ & $-0.61$ & ns\\
\bottomrule\end{tabular}
\caption{Spearman correlations of spectral metrics with head-type scores (GPT-2, $n{=}144$; all shown are
FDR-significant). Copying is an orthogonal OV-side control.}
\label{tab:taxonomy}
\end{table}

\section{Causal anatomy on GPT-2: \texorpdfstring{$\MS\gg\MA$}{MS >> MA}}\label{sec:causal}
\textbf{Symmetrization ablation.} Replacing a head's \M with \MS (killing \MA) or \MA (killing \MS) and measuring
held-out CE, the \textbf{symmetric part is $6.7\times$ more load-bearing}: $\sum \Delta\mathrm{CE}_{\text{anti}}{=}2.60$
vs $\sum\Delta\mathrm{CE}_{\text{sym}}{=}0.39$; killing \MA costs $\le0.02$ general CE anywhere. But a \emph{targeted
induction task} exposes that \MA is causally \emph{essential} for specific heads: killing it on the canonical induction
head~5.1 raises induction loss $8\times$ ($0.21\to1.77$; $15\times$ the next-largest effect) and on the previous-token
head~4.11 by ${\sim}50\%$. Canonical induction heads 5.5/6.9 show smaller positive effects ($+0.03$, ranks 7--8 of
144); the top-5 by effect are 5.1, 4.11, 6.7, 3.0, 8.5, so top-5 overlap with the canonical circuit is 2/5
(hypergeometric $p{=}0.014$), with the argmax landing on a canonical induction head --- circuit-level enrichment, not
an exact circuit match. Crucially this importance is a \textbf{discrete circuit fact}: no spectral scalar (\dirfrac,
\Dhead, \imagmass, $\|\MA\|$) predicts \emph{which} heads need \MA (all $|\rho|\lesssim0.1$), so the 144-head
aggregate is null.

\textbf{Incremental validity.} For predicting the previous-token score, the complex refinement adds no reliable
positive increment on GPT-2: $\mathrm{partial}\,\rho(\Dhead,\text{prev}\mid\dirfrac)=-0.21$ (layer-clustered 95\% CI
$[-0.51,+0.08]$ --- the point estimate is negative but not reliably so; among prev-candidate heads,
$\text{prev}{>}0.05$, the residual association is strongly negative, $-0.63$), and in a nested regression on the
ablation effect \dirfrac explains $0.001$ of variance rising only to $0.028$ with \Dhead added, the added term
wrong-signed. On learned-absolute positions, the plain antisymmetric magnitude captures everything the complex
spectrum reliably offers about its strong prev heads; \S\ref{sec:reversal} shows this head-profile pattern
generalizes to OPT-1.3B and GPT-Neo-1.3B, though their aggregate statistics do not.

\section{The architecture-conditional signature (main result)}\label{sec:reversal}
\begin{figure}[t]\centering
\includegraphics[width=\linewidth]{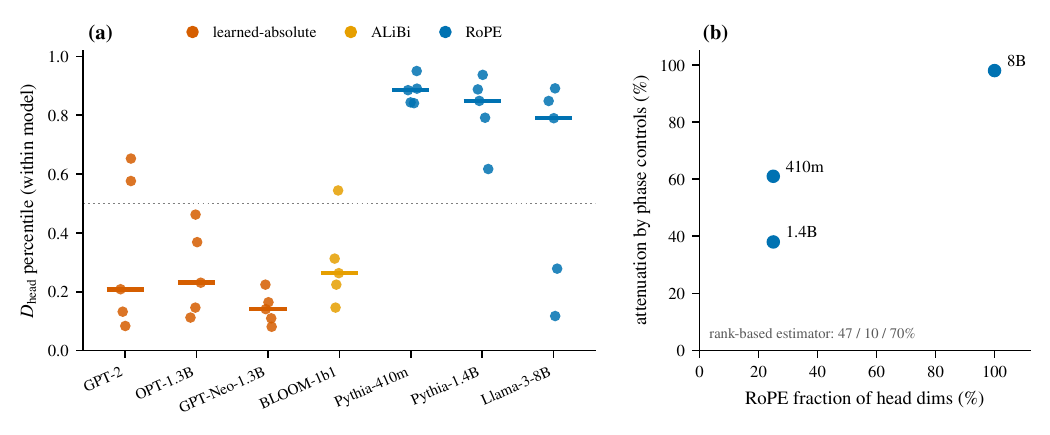}
\caption{\textbf{(a)} Within-model \Dhead percentile of each model's top-5 previous-token heads: bottom-quartile
under learned-absolute (red) and ALiBi (orange) positions, top-quintile under \Rope (blue); joint
Mann--Whitney $p{=}2.5{\times}10^{-5}$; bars mark per-model medians. \textbf{(b)} Attenuation of \Dhead's
advantage by \Rope-phase controls grows with \Rope fraction (hybrid estimator: $61\%/38\%$ at $25\%$, ${\sim}98\%$ at
$100\%$; see Table~\ref{tab:mediation} for estimator sensitivity --- with two dose levels this is suggestive, not an
established dose--response).}
\label{fig:main}
\end{figure}

We run the incremental-validity test across positional schemes on seven models (Table~\ref{tab:reversal}). On every
\Rope model, \Dhead's incremental contribution beyond \dirfrac is positive and robust (clustered CIs
$[+0.21,+0.59]$). Under learned-absolute positions, however, the \emph{aggregate} statistic is heterogeneous
(GPT-2 $-0.21$; OPT-1.3B $+0.35$; GPT-Neo-1.3B $+0.20$) --- and decomposing it shows why it is the wrong statistic: a
positive gradation among near-zero-prev \emph{bulk} heads is present in most models (including GPT-2: $+0.25$ within
$\text{prev}{\le}0.05$), while the signal that differs by architecture lives in the \emph{strong} previous-token
heads themselves. The last column of Table~\ref{tab:reversal} gives each model's top-5 prev heads' within-model
\Dhead percentile: \textbf{bottom-quartile under learned absolute} (medians $0.21/0.23/0.14$; e.g.\ GPT-2's 4.11 at
$0.22$; all with content-like symmetric profiles, \contentpos $0.46$--$0.57$) and ALiBi ($0.26$) versus \textbf{top-quintile under
\Rope} (medians $0.89/0.85/0.79$; Fig.~\ref{fig:main}a). Holding
$d_k{=}64$ fixed (GPT-2, OPT-1.3B, Pythia-410m: $0.21/0.23$ vs $0.89$) ties the contrast to the positional scheme,
not head dimension.

\textbf{Model-level inference and sensitivity.} Heads within a model share training, layers, and detectors, so the
model is the honest experimental unit. At model level the separation is \emph{perfect} --- every non-\Rope model's
top-5 median sits below every \Rope model's --- giving an exact model-level permutation $p{=}1/\binom{7}{3}{=}0.029$,
the minimum attainable with seven models; the head-level joint Mann--Whitney $p{=}2.5{\times}10^{-5}$ is reported as
descriptive only. The contrast is not an artifact of the $k{=}5$ choice: model-level separation is perfect at every
$k\in\{1,3,5,10\}$, and leave-one-model-out preserves it for all seven models. The binding statistical limit is the
model count itself --- one ALiBi and one full-\Rope family --- which more model families, not more heads, would
address (Limitations). The defensible claim is \emph{profile-level}: \textbf{the heads that implement previous-token
attention are spectrally rotational under \Rope and spectrally non-rotational (content-like) under learned absolute
positions.} Mechanistically, this is what the positional algebra dictates: with learned absolute embeddings,
prev-token attention can be built by \emph{matching} adjacent position vectors --- a symmetric, content-like
operation --- whereas \Rope builds it from \emph{rotations}, imprinting complex eigenstructure. \textbf{ALiBi
provides the cleanest test}: position enters as a score \emph{bias}, so QK needs no positional structure at all ---
and BLOOM-1b1's top prev heads are indeed non-rotational (percentile $0.26$), patterning with the learned-absolute
group exactly as the mechanism predicts. Under \Rope these
heads sit at the \dirfrac null (${\sim}0.70$) with \Dhead at the Ginibre value (${\approx}0.61$) while the
population suppresses below it (medians $0.40$--$0.45$; per-model matched nulls: 410m prev heads' \Dhead
$z\in[-0.1,+5.0]$ vs population median $z{=}{-}11$; 1.4B prev-head median $z{=}{-}3.7$ vs population
$z{=}{-}13.9$): they \emph{retain} near-random rotational structure that other heads learn to suppress, and the
Schur/complex view detects this where the Frobenius magnitude cannot.

\textbf{Robustness and scope.} Three qualifications sharpen the claim. (i)~\emph{Metric specificity}: among
complex-spectrum summaries, \imagmass is incrementally positive on all models and Henrici non-normality negative on
all; \Dhead composites these two channels. No single aggregate scalar separates the schemes --- the head-profile
statistic does. (ii)~\emph{Class vs bulk}: the \Rope positives are class-level separation (rank-biserial
$+0.44/+0.60$ at $\text{prev}{>}0.2$ in Pythia; $+0.29$ at $\text{prev}{>}0.1$ in Llama) \emph{plus} bulk gradation;
within confirmed prev heads there is no further gradation (subset partials n.s.). Under learned-absolute positions
the class-level tests are null-to-negative at the strong end (GPT-2 $-0.39$ at $\text{prev}{>}0.2$, significant;
OPT $-0.11$ n.s.; GPT-Neo $+0.13$ n.s.), though GPT-Neo shows a positive signal at the looser $\text{prev}{>}0.1$
band ($+0.34$) driven by its moderate-prev heads --- its \emph{top} prev heads still sit at percentile $0.14$. The
learned-absolute models are thus heterogeneous in their moderate bands (candidate sources: training corpus,
GPT-Neo's alternating local-attention layers) while uniform at the top of the class. (iii)~GPT-2's negative
aggregate concentrates exactly among prev-candidate heads ($-0.63$ at $\text{prev}{>}0.05$); OPT's positive
aggregate is entirely bulk ($-0.18$ among its prev candidates). Aggregate partials should not be read as the
architecture signal in either direction.

\begin{table}[t]\centering\footnotesize\setlength{\tabcolsep}{3.5pt}
\begin{tabular}{lcccccccc}\toprule
model & $d_k$ & positions & norm & $\rho_{\texttt{dir}}$ & $\rho_{D}$ & partial & 95\% CI & \textbf{top-5 pctile}\\\midrule
GPT-2        & 64  & absolute   & LN  & $+0.47$ & $+0.29$ & $-0.21$ & $[-0.51,+0.08]$ & $\mathbf{0.21}$\\
OPT-1.3B     & 64  & absolute   & LN  & $+0.08$ & $+0.28$ & $+0.35$ & $[+0.19,+0.48]$ & $\mathbf{0.23}$\\
GPT-Neo-1.3B & 128 & absolute   & LN  & $+0.35$ & $+0.41$ & $+0.20$ & $[+0.07,+0.33]$ & $\mathbf{0.14}$\\
BLOOM-1b1    & 96  & ALiBi      & LN  & $+0.13$ & $+0.25$ & $+0.34$ & $[+0.22,+0.47]$ & $\mathbf{0.26}$\\
Pythia-410m  & 64  & \Rope 25\% & LN  & $+0.57$ & $+0.62$ & $+0.35$ & $[+0.21,+0.47]$ & $\mathbf{0.89}$\\
Pythia-1.4B  & 128 & \Rope 25\% & LN  & $+0.31$ & $+0.60$ & $+0.50$ & $[+0.38,+0.59]$ & $\mathbf{0.85}$\\
Llama-3-8B   & 128 & \Rope 100\%& RMS & $+0.32$ & $+0.38$ & $+0.33$ & $[+0.21,+0.45]$ & $\mathbf{0.79}$\\
\bottomrule\end{tabular}
\caption{Incremental validity of \Dhead over \dirfrac for the previous-token score, seven models / three
positional schemes. $\rho_{\texttt{dir}}=\rho(\dirfrac,\text{prev})$, $\rho_D=\rho(\Dhead,\text{prev})$; partial $=
\mathrm{partial}\,\rho(\Dhead,\text{prev}\mid\dirfrac)$ (Spearman of OLS residuals); CIs: layer-clustered bootstrap
(heads share layers/training, design effects $3.6$--$6.9$; Llama heads additionally share K-projections within GQA
groups of 4). \textbf{The aggregate partial is heterogeneous under learned-absolute positions} ($-0.21/+0.35/+0.20$)
because it mixes a near-ubiquitous positive bulk gradient (among near-zero-prev heads) with the class-level signal;
\textbf{the architecture-conditional statistic is the last column} --- the median within-model \Dhead percentile of
the top-5 previous-token heads: bottom-quartile under learned-absolute and ALiBi vs top-quintile under \Rope (perfect model-level separation,
exact permutation $p{=}0.029$; head-level $p{=}2.5{\times}10^{-5}$, descriptive). $d_k{=}64$ is shared by GPT-2, OPT-1.3B, and Pythia-410m, so the contrast
tracks the positional scheme, not head dimension. All \Rope partials survive BH-FDR.}
\label{tab:reversal}
\end{table}

\section{Mechanism: \texorpdfstring{\Dhead}{Dhead} tracks the RoPE phase}\label{sec:mechanism}
\Rope pairs each head's $d_k$ dims (rotate-half convention, verified) into frequencies $\theta_t$; the QK interaction
per frequency is a rank-1 complex sub-operator $\Mt=w_q^t\,\overline{w_k^t}^{\top}$, with the static
$\M=\sum_t\mathrm{Re}(\Mt)$. The \emph{directional} (relative-position-asymmetric) content is carried by the imaginary
part:
\begin{equation}
\mathrm{score}(\Delta)-\mathrm{score}(-\Delta)=\sum_t 2\sin(\Delta\theta_t)\,n^\top \mathrm{Im}(\Mt)\, n .
\end{equation}
(Full-score reconstruction from $\{\Mt,\theta_t\}$ matches the model to $10^{-6}/10^{-2}$ in fp32/bf16.) Empirically,
previous-token heads have high rotary directional fraction \ropeimag (correlation with prev: $+0.53/+0.42/+0.50$ for
410m/1.4B/Llama) concentrated at \emph{high frequencies} ($+0.68$ on full-\Rope Llama---a
\citet{barbero2025rope,urrutia2025decoupling} replication), and \textbf{\Dhead tracks this deployment}. Adding per-frequency \Rope-phase summaries as controls \emph{attenuates}
\Dhead's residual association substantially (Table~\ref{tab:mediation}) --- most completely on full-\Rope Llama,
where the default-estimator residual is indistinguishable from zero ($+0.006$, $p{=}0.84$; fully rank-based:
$+0.075$, $p{=}0.016$). Two qualifications. First, all quantities are co-derived from the same weights, so this is
shared-variance accounting, not causal mediation. Second, the attenuating covariate \emph{changes identity} with
\Rope fraction: on partial-\Rope Pythia the attenuation runs through \ropeimag (how \emph{much} rotary content is
directional; \texttt{freq\_centroid} alone attenuates nothing, $+0.35$), whereas on full-\Rope Llama it runs almost
entirely through \texttt{freq\_centroid} (\emph{where in frequency} the directional mass sits: alone
$+0.33\to-0.00$, while \ropeimag alone leaves $+0.32$) --- consistent with all directional content being rotary by
construction at $100\%$, shifting the informative axis from amount to location. The pattern is consistent with
fuller attenuation at higher \Rope fraction, but with two dose levels and within-dose spread ($61\%$ vs $38\%$)
comparable to the dose gap, we do not press a dose--response reading. The relative-position kernel confirms these
heads peak at $\Delta{=}{-}1$ with \Rope-periodic oscillation (Fig.~\ref{fig:kernels}, left).

\textbf{Placebo controls.} Two controls calibrate what the attenuation means. \emph{(i) \Rope-model specificity:} on
all three learned-absolute models (no \Rope), computing a pseudo-\ropeimag with a fake rotate-half pairing over all
head dims yields no prev signal (GPT-2 $-0.14$, OPT-1.3B $-0.06$, GPT-Neo-1.3B $-0.09$; all n.s., vs
$+0.42$--$+0.53$ on \Rope models), and adding the pseudo-phase features barely attenuates ($4$--$22\%$): the recipe
finds nothing where \Rope is absent. \emph{(ii) Pairing
specificity:} randomly permuting the rotary columns of $W_Q,W_K$ --- which leaves \M (hence \Dhead, \dirfrac)
unchanged and preserves rotary-\emph{block} membership, but scrambles the pairing and frequency assignment --- still
attenuates $31$--$38\%$ on average, so much of the attenuation is carried by rotary-block membership alone. The
increment specific to the \emph{true} pairing/frequency assignment is marginal on Pythia-410m (true $60.7\%$ vs the
permutation null, empirical $p{=}0.039$, $n{=}50$ permutations) and not significant on Pythia-1.4B ($37.7\%$,
$p{=}0.12$). The mechanism is thus \Rope-specific and largely block-level; the finer pairing-level reading should be
held lightly.

\begin{table}[t]\centering\footnotesize\setlength{\tabcolsep}{4pt}
\begin{tabular}{lccccc}\toprule
model & \Rope frac. & partial & $+$phase controls & atten.\ (hybrid) & atten.\ (rank)\\\midrule
Pythia-410m & $25\%$  & $+0.35$ & $+0.14$ & $61\%$ & $47\%$\\
Pythia-1.4B & $25\%$  & $+0.50$ & $+0.31$ & $38\%$ & $10\%$\\
Llama-3-8B  & $100\%$ & $+0.33$ & $+0.006$ ($p{=}0.84$) & ${\sim}98\%$ & $70\%$\\
\bottomrule\end{tabular}
\caption{Attenuation of \Dhead's residual prev-token association (partial $=\mathrm{partial}\,\rho(\Dhead,
\text{prev}\mid\dirfrac)$) when per-frequency \Rope-phase summaries
(\ropeimag, \texttt{freq\_centroid}) are added as controls. ``Hybrid'' is the paper's default estimator (Spearman of
OLS residuals); ``rank'' is fully rank-based, under which the Llama residual is small but nonzero ($+0.075$,
$p{=}0.016$). All quantities are deterministic functions of the same weights: this is shared-variance accounting
among co-derived features, \emph{not} causal mediation.}
\label{tab:mediation}
\end{table}

\section{Causal confirmation: ablating \texorpdfstring{$\mathrm{Im}(\Mt)$}{Im(Mt)}}\label{sec:ablation}
\begin{figure}[t]\centering
\includegraphics[width=0.49\linewidth]{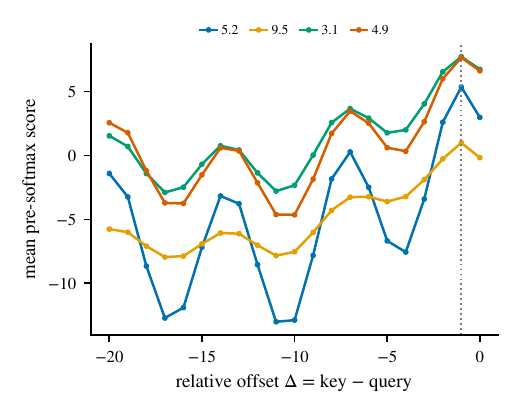}\hfill
\includegraphics[width=0.49\linewidth]{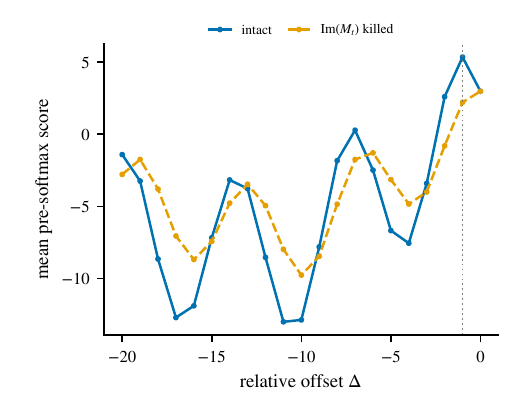}
\caption{\textbf{Left:} corpus-averaged relative-position score kernels of top previous-token heads (Pythia-410m) peak
at $\Delta{=}{-}1$ with \Rope-periodic oscillation. \textbf{Right:} ablating $\mathrm{Im}(\Mt)$ on the induction-feeding
head~5.2 flattens the $\Delta{=}{-}1$ peak (and raises induction loss $4\times$).}
\label{fig:kernels}
\end{figure}

We ablate a head's directional \Rope content by zeroing $\mathrm{Im}(\Mt)$ (dropping the $\sin(\Delta\theta)$
component), which \textbf{symmetrizes} its relative-position kernel; the decomposition
$\text{full}=\text{symmetric}+\text{directional}$ is verified to $6\times10^{-6}$ (fp32; $2\times10^{-5}$ on bf16
Llama). Sweeping the ablation over \emph{all} heads of all three \Rope models, the largest induction-loss effect
lands \textbf{on a top-2 prev-scoring head every time} --- an identity predicted a priori: Pythia-410m head~5.2
(prev rank 1/384; prev~$0.95$, $\Dhead\,0.61$): induction CE $0.52\to2.24$ (${\sim}4\times$; exact permutation
$p{=}1/384$); Pythia-1.4B head~1.11 (rank 1/384): $0.56\to1.11$ (${\sim}2\times$; $p{=}1/384$); Llama-3-8B
head~L14.H26 (rank 2/1024; prev~$0.65$, $\Dhead\,0.61$): $0.30\to0.91$ (${\sim}3\times$; $p{=}2/1024$) --- jointly
${\approx}10^{-8}$ under random assignment. (Llama's \#1 prev head sits in layer~0, too early to feed induction;
the causally-critical head is the mid-depth prev head.) Effects are highly concentrated: $92$--$99\%$ of heads move
by $|\Delta\text{Ind}|<0.01$; the \emph{other} $\text{prev}{>}0.5$ heads show ${\approx}0$ effects (consistent with
redundancy across a multi-head circuit), and a few low-prev heads show small effects ($\le0.17$), suggesting
downstream circuit participation the prev detector does not capture; head-level aggregate correlations are
correspondingly null. The before/after kernel shows the $\Delta{=}{-}1$ peak flattening
(Fig.~\ref{fig:kernels}, right). As in \S\ref{sec:causal}, causal importance is concentrated, not a smooth spectral
gradient; what the ablation establishes is an \emph{existence-with-predicted-identity} claim: the rotational phase
\Dhead reads is causally load-bearing for positional routing in the specific heads that carry it.

\section{When the imprint arises: a checkpoint natural history}\label{sec:dynamics}
The static profile (\S\ref{sec:reversal}) is an end-state. To see how it arises we track 22 public checkpoints each
of Pythia-410m and Pythia-160m (log-spaced steps $0\ldots143{,}000$; the dense early grid straddles the induction
onset at ${\sim}2{\times}10^9$ tokens \citep{tigges2024llm}), measuring per head: the audited weight metrics
(\dirfrac, \Dhead, \ropeimag), behavioral previous-token and prefix-matching scores, an in-context-learning proxy
(second-copy minus first-copy NLL on repeated-random sequences), and the K-composition wiring between each model's
eventual prev head and its induction heads. (The ``eventual prev head'' is identified retrospectively at the
final checkpoint; Q3 below tests --- and rejects --- its prospective identifiability.)

\begin{figure}[t]\centering
\includegraphics[width=\linewidth]{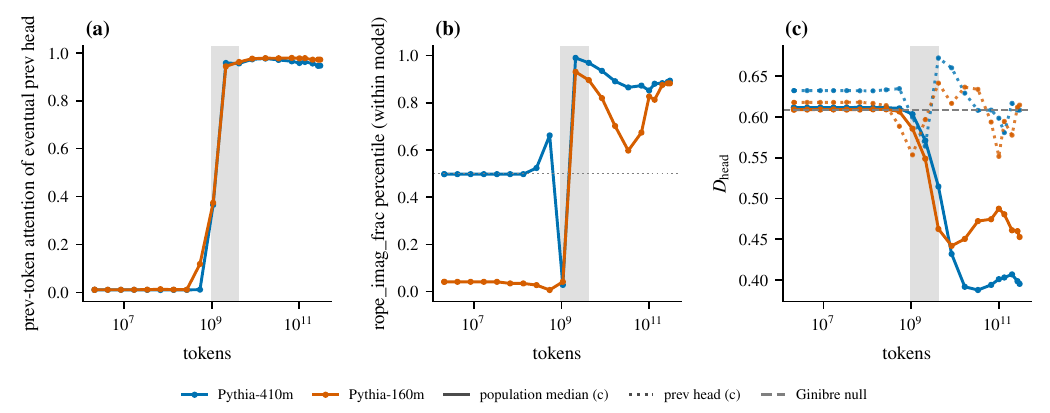}
\caption{Checkpoint natural history, Pythia-410m (blue) and 160m (red); grey band = the formation window
($1$--$4\times10^9$ tokens). \textbf{(a)} previous-token behavior forms sharply and identically in both models.
\textbf{(b)} the prev head's \ropeimag\ percentile locks in \emph{with} behavior, not before. \textbf{(c)} the
population median \Dhead is suppressed below the Ginibre null only \emph{after} formation.}
\label{fig:dynamics}
\end{figure}

\textbf{A three-act history at half-doubling checkpoint resolution, replicated across both models}
(Fig.~\ref{fig:dynamics}).
\emph{Act I (0--0.5B tokens): silence at the null.} Every head sits at the Ginibre null (behavior $0.01$;
\ropeimag $0.500$; population median \Dhead $0.61$; K-composition at baseline) --- step 0 doubles as an in-vivo
verification of the paper's random-orientation null. \emph{Act II (1--4B tokens): sharp formation.} Previous-token
behavior jumps first ($0.37$ at step 512 $\to$ $0.95$ at step 1000 --- the same steps, and even the same $0.37$
waypoint, in both models), with induction behavior, the ICL proxy, and the K-composition wiring following within the
same window. The prev head's \emph{within-model rotary-phase (\ropeimag) percentile} snaps from at-or-below chance ($0.03$--$0.04$
at step 512, mid-formation, in both models) to the top decile ($0.90$--$0.99$) by step 1000: the signature locks in
\emph{with} the behavior --- at our half-doubling resolution, neither leads. \emph{Act III (4--300B tokens): slow
differential sculpting.} The population median \Dhead is suppressed well below the null ($0.61\to0.395$ / $0.453$)
over the next $100\times$ of training while the previous-token heads retain it, and the absolute rotary-phase content of
the prev head consolidates slowly ($0.500\to0.578$, same endpoint in both models). The static ``retain-vs-suppress''
profile of \S\ref{sec:reversal} is therefore the end state of a post-formation differentiation process.

\textbf{Pre-registered answers.} Q1 (lead/lag): the percentile signature is \emph{simultaneous} with formation; the
absolute consolidation \emph{lags}. Q2: population suppression is a post-formation process. Q3 (predictive
signature): \textbf{no} --- before formation the eventual prev head is not reliably identifiable from its spectrum
(its percentile even sits low mid-formation); we report this pre-registered negative plainly. The natural history
thus bounds what statics can claim: the rotational imprint is the fingerprint the algorithm leaves in weight space,
consolidated after function --- and because observation cannot order structure and function within the formation
window, necessity can only be settled by intervention.

\section{Is the rotational channel necessary? Constrained-training interventions}\label{sec:intervention}
We train 2-layer attention-only models (d${=}128$, 4 heads, $d_k{=}32$) from scratch on a per-sequence random-map
task --- $x_{t+1}=f_{\mathrm{seq}}(x_t)$ w.p.\ $0.9$, uniform noise w.p.\ $0.1$, with $f_{\mathrm{seq}}$ resampled
every sequence --- which forces induction (no global memorization; no fixed-offset shortcut) and yields a crisp
formation time. Grid: $\{$APE, \Rope$\}\times\{$free; sym-\M: penalize $\|\MA\|_F^2/\|\M\|_F^2$;
$\mathrm{Im}(\Mt)$-suppressed (\Rope only): penalize $\sum_t\|\mathrm{Im}(\Mt)\|_F^2/\sum_t(\|\mathrm{Re}\|^2{+}
\|\mathrm{Im}\|^2)\}$, $n{=}5$ seeds. A load-bearing algebraic point shapes the grid: \textbf{symmetrizing the
static \M does not zero $\mathrm{Im}(\Mt)$} (the static operator contains only the $\mathrm{Re}$ parts,
$\M=\sum_t\mathrm{Re}(\Mt)+\M_{\text{non-rot}}$), so the two constraints dissociate the static-antisymmetry and
phase channels. Both penalties enforce hard (final $\mathrm{Im}$ share $0.000$; \dirfrac $0.004$--$0.006$) and no
arm loses final capability (all reach the task floor with strong prev and induction heads).

\begin{figure}[t]\centering
\includegraphics[width=\linewidth]{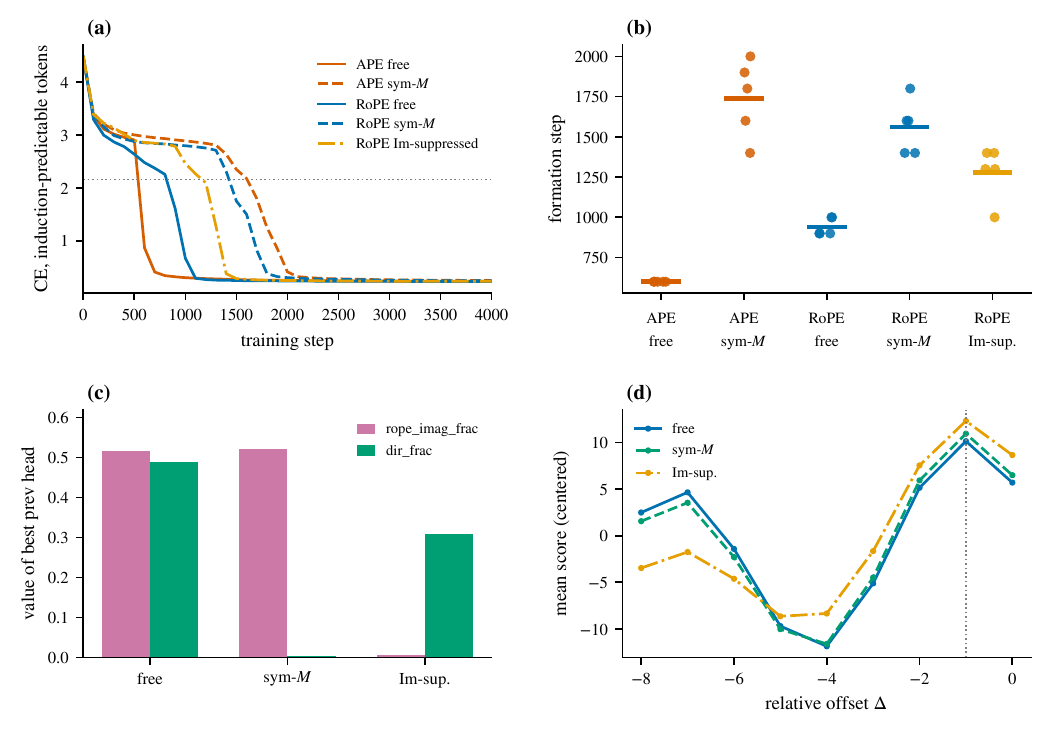}
\caption{Constrained-training interventions. \textbf{(a)} induction-capability curves: every arm reaches the floor;
constraints delay the drop. \textbf{(b)} formation times ($n{=}5$). \textbf{(c)} the dissociation: under sym-\M the
prev head keeps full phase content (\ropeimag) with a fully symmetric static \M (\dirfrac${\approx}0$); under
Im-suppression the reverse. \textbf{(d)} positional kernels: sym-\M is indistinguishable from free (the kernel is
carried by the untouched phase); Im-suppression reshapes the kernel but its peak stays at $\Delta{=}{-}1$
(reroute).}
\label{fig:intervention}
\end{figure}

\textbf{Results against the pre-registered predictions} (Fig.~\ref{fig:intervention};
Table~\ref{tab:intervention}). P1 (free base rates) holds: APE forms fastest ($600\pm0$ steps), \Rope free at
$940\pm55$. P2's strong form is \textbf{falsified}: suppressing the phase to zero delays formation only
${\sim}1.4\times$ and the circuit \emph{reroutes} --- the prev head re-forms with $\ropeimag{=}0.005$ and a
qualitatively different (cos-only) relative-position kernel whose peak still lands at $\Delta{=}{-}1$. P3 is
\textbf{falsified in the opposite direction, our largest surprise}: forcing symmetry on APE is the \emph{costliest}
constraint in the grid ($600\to1740\pm241$, $2.9\times$) --- the fast APE solution is \emph{antisymmetric}
embedding-matching, and the symmetric variant (nearby-kernel $+$ causal mask), while reachable, is much harder to
find; trained learned-absolute LLMs end at symmetric profiles (\S\ref{sec:reversal}) with $1000\times$ more data,
and the toy exposes the search cost of that end state. P4's formation-time version fails (RoPE+sym delays
$1.66\times$) but its \textbf{mechanistic dissociation lands exactly}: the sym-arm's prev head carries
\emph{full} phase content ($\ropeimag\ 0.52$) through a \emph{fully symmetric} static operator
($\dirfrac\ 0.004$--$0.005$), and its positional kernel is indistinguishable from the free arm's --- directional
attention through a symmetric \M, which is impossible under absolute positions.

\begin{table}[t]\centering\small
\begin{tabular}{lcccc}\toprule
pre-registered contrast (formation delay) & MWU exact $p$ & $q_{\mathrm{BH}}$ & rank-biserial & HL shift\\\midrule
P1\ \ \Rope free $>$ APE free          & $0.0040$ & $0.016$ & $+1.00$ & $+300$ steps\\
P3\ \ APE sym-\M $>$ APE free          & $0.0040$ & $0.008$ & $+1.00$ & $\mathbf{+1200}$ steps ($2.9\times$)\\
P4\ \ \Rope sym-\M $>$ \Rope free      & $0.0040$ & $0.005$ & $+1.00$ & $+600$ steps\\
P2\ \ \Rope Im-sup $>$ \Rope free      & $0.0079$ & $0.008$ & $+0.92$ & $+400$ steps\\
\bottomrule\end{tabular}
\caption{Constrained-training formation delays ($n{=}5$ seeds per cell; one-sided exact Mann--Whitney, BH-FDR over
the four pre-registered contrasts). Every constraint delays formation significantly; none blocks it.}
\label{tab:intervention}
\end{table}

\textbf{Verdict.} No spectral channel is \emph{necessary} --- the solution space is degenerate and training reroutes
around every constraint we imposed --- but each constraint carries a significant, quantifiable search cost
(Table~\ref{tab:intervention}), and the cost structure identifies each scheme's \emph{default} solution. With
\S\ref{sec:causal} and \S\ref{sec:ablation} this completes a three-way distinction the field often conflates:
\emph{trained-circuit dependence} (post-hoc ablations are catastrophic), \emph{developmental preference} (defaults
are found much faster), and \emph{necessity} (nothing here is strictly necessary). A companion note
\citep{li2026steering} tests the constructive converse --- ban-free \emph{assistance} --- and finds it selects the
same implementations far more cheaply than banning them ($1.3\times$ vs.\ $2.9\times$ for the antisymmetric-to-symmetric
flip) while solution-specific initialization accelerates formation outright, evidence that this spectral economy is
not only priced but steerable.

\section{Discussion}
\textbf{What earns its keep, and where.} The plain symmetric/antisymmetric split of \M is an architecture-general
descriptor of head function (\S\ref{sec:descriptive}) and, causally, the symmetric part is the workhorse
(\S\ref{sec:causal}). The \emph{complex-eigenvalue refinement} is architecture-conditional at head level: the previous-token solution is
spectrally non-rotational (content-like) under learned-absolute positions and rotational under \Rope, where \Dhead
tracks the rotational phase whose causal load \S\ref{sec:ablation} establishes head-locally
(\S\ref{sec:reversal}--\ref{sec:ablation}). This gives the recent ``non-Hermitian transformer'' program a concrete,
controlled home---the complex structure of the \emph{QK operator} does measurable, causal work precisely when position
is encoded as complex phase---while honestly bounding it: on GPT-2 the non-Hermitian label is decorative.

\textbf{Practical reading.} For weight-only head triage: under \Rope, \ropeimag\ (or the \Dhead percentile) flags
positional-routing heads without a forward pass; under absolute/ALiBi schemes the same statistics are
\emph{misleading in aggregate} (bulk gradients) and only the strong-head profile is informative. For
interpretability methodology, the dynamics result is a caution: a clean static weight--function correspondence
need not be predictive during training, and post-hoc ablation severity must not be read as developmental necessity.

\textbf{Companion work (context only).} Companion work in preparation analyzes the \emph{iterated} attention
propagator with pseudospectral tools under mask-structure nulls; none of this paper's claims relies on it. We flag
two of its directions as context: a division of labor consistent with our tool choice (eigenvalue-level summaries
suffice for the static scoring form; resolvent-level tools are needed for the depth-iterated propagator), and a
state-dependent character of induction heads that static weight fingerprints cannot capture --- consistent with
their content-like static profile in \S\ref{sec:descriptive}.

\textbf{Limitations.} (i)~The \Rope evidence in \S\ref{sec:reversal}--\ref{sec:mechanism} is correlational
(shared-variance) over heads; \S\ref{sec:ablation} supplies causation but as a concentrated per-head effect (now on
all three \Rope models). (ii)~Three learned-absolute models are tested with a consistent strong-head profile but heterogeneous aggregate and
moderate-band statistics, whose sources (training corpus; GPT-Neo's alternating local attention) are unresolved; the ALiBi scheme is covered
by a single model (BLOOM-1b1). (iii)~Inference treats heads as units: heads share layers and training (layer-clustered
design effects $3.6$--$6.9$; we report clustered CIs), and Llama heads share K-projections within GQA groups of 4
(between-group variance $0.55$--$0.79$) --- group-level dependence is only partially resolved. (iv)~Llama runs in
bf16 (float64 metrics from bf16 weights are exact w.r.t.\ the model; the diagonal convention check matches to bf16
precision only). (v)~Single-head ablations miss composition; the prev$\to$induction wiring is verified by K-composition
(\S\ref{sec:descriptive}) but full path patching is future work.
(vi)~QK biases lie outside \M and are held fixed in ablations; the prev-detector's bulk-ordering reliability is
unquantified. (vii)~A non-Llama full-\Rope family
(Qwen/Mistral) would broaden the third architecture point.

\section{Conclusion}
For the attention QK operator, the plain antisymmetric split describes head function architecture-generally; the
complex-eigenvalue refinement earns its keep only under \Rope, where it tracks the rotational phase whose causal
load head-level ablations establish. Tracking training shows this spectral signature is born at the random-matrix null, locks in with circuit
formation, and is sculpted into today's profiles after function arrives; constrained training shows it is a
\emph{default}, not a necessity --- every spectral channel we blocked was rerouted around, at a significant and
interpretable search cost. The non-Hermitian view of attention is neither uniformly profound nor uniformly
decorative: \emph{the positional scheme sets the default spectral algebra of attention's solutions}, and we
measure that claim statically, dynamically, and causally in the settings analyzed. Scaled constrained pretraining
is its direct falsification path.

\paragraph{Reproducibility \& pre-registration.}\sloppy Code in \texttt{src/}: \texttt{extract}, \texttt{decompose},
\texttt{metrics}, \texttt{observables}, \texttt{semantics}, \texttt{p4\_rope}, \texttt{rope},
\texttt{p5\_rope\_ablation}, \texttt{p9\_checkpoint\_dynamics}, \texttt{p10\_training\_intervention}. Per-head
tables: \texttt{results/cache/*\_head\_full.parquet}; checkpoint trajectories and training histories under
\texttt{results/cache/}. The dynamics questions (Q1--Q3) and intervention predictions (P1--P4) were pre-registered
in the project plan before data collection; falsified predictions (Q3, P2-strong, P3, P4-timing) are reported as
such. The full artifact bundle --- code, per-head tables, checkpoint trajectories, training histories, figure
scripts, and the pre-registration documents --- is released at
\url{https://github.com/HengyuLi-Ozaki-lab/qk-spectral-fingerprint}; the repository's version history stamps
which analyses were specified before their data existed.

\bibliographystyle{plainnat}
\bibliography{references}

\end{document}